\documentclass[a4paper]{article}

\usepackage{INTERSPEECH2016}

\usepackage{color}
\usepackage{graphicx}
\usepackage{amsmath}
\usepackage{amsfonts}
\usepackage{multirow}
\newcommand{\tail}{\text{tail}}
\newcommand{\head}{\text{head}}
\newcommand{\term}[1]{\textbf{#1}}
\newcommand{\adj}{\text{adj}}
\DeclareMathOperator*{\argmax}{\arg\!\max}
\usepackage{tikz}

\newcommand{\klcomment}[1]{\textcolor{green}{\bf \small [ #1 --KL]}}
\newcommand{\kpgcomment}[1]{\textcolor{blue}{\bf \small [ #1 --KPG]}}
\newcommand{\htcomment}[1]{\textcolor{red}{\bf \small [ #1 --HT]}}
\renewcommand{\klcomment}[1]{\ignorespaces}
\renewcommand{\kpgcomment}[1]{\ignorespaces}
\renewcommand{\htcomment}[1]{\ignorespaces}

\ninept

\title{Efficient Segmental Cascades for Speech Recognition}

\makeatletter
\def\name#1{\gdef\@name{#1\\}}
\makeatother \name{{\em Hao Tang, Weiran Wang, Kevin Gimpel, Karen Livescu}}

\address{Toyota Technological Institute at Chicago \\
  {\small \tt \{haotang,wwang5,kgimpel,klivescu\}@ttic.edu}
}

\begin{document}

\maketitle

\begin{abstract}
Discriminative segmental models offer a way to incorporate flexible
feature functions into speech recognition.  However, their appeal has
been limited by their computational requirements, due to the large
number of possible segments to consider.  Multi-pass cascades of
segmental models introduce features of increasing complexity in
different passes, where in each pass a segmental model rescores
lattices produced by a previous (simpler) segmental model.  In this
paper, we explore several ways of making segmental cascades efficient
and practical:  reducing the feature set in the first pass, frame
subsampling, and various pruning approaches.  In experiments on
phonetic recognition, we find that with a combination of such
techniques, it is possible to maintain competitive performance while
greatly reducing decoding, pruning, and training time.
\end{abstract}
\noindent{\bf Index Terms}: segmental conditional random fields,
discriminative segmental cascades, structure prediction cascades,
multi-pass decoding

\section{Introduction}

Segmental models have been proposed for automatic speech recognition
(ASR) as a way to
enable ASR systems to use informative features at the segment level,
such as duration, trajectory information, and segment-level
classifiers~\cite{Glass2003, ZweigNguyen2009}.
Discriminative segmental models
such as segmental conditional random fields (SCRFs) \cite{ZweigNguyen2009} and segmental
structured support vector machines (SSVMs)~\cite{ZhangEtAl2010}
allow for flexible integration of a variety of such information
sources via
feature functions with learned weights.

One of the main impediments to the broad use of segmental models is
their computational requirements.  Since, in principle, each edge in
the graph of potential segmentations must be considered separately,
segmental models involve a great deal more computation than do
frame-based models.  This challenge has made it difficult to apply
segmental models in first-pass decoding, and most segmental ASR work
has been done by rescoring lattices produced by first-pass frame-based
systems~\cite{ZweigNguyen2009, ZweigEtAl2011, WachterEtAl2007}.
In recent work on phonetic
recognition, it has been shown that
it is possible to achieve high-quality first-pass recognition with
segmental models~\cite{AbdelHamidEtAl2013,He2015,LuEtAl2016}, 
but computation is still a
concern.  It is possible to obtain large speed improvements through
cleverly constrained feature functions~\cite{HeFoslerLussier2012}, but in
the general case of arbitrary segmental feature functions the problem
remains.

In recent work we introduced discriminative segmental cascades (DSC)~\cite{TangEtAl2015}, 
a multi-pass approach in which each pass is a segmental
model.  This avoids any reliance on frame-based systems to produce
lattices and allows the use of arbitrary segmental feature functions,
while still producing roughly real-time decoding for phonetic recognition.  This is
achieved by delaying the more expensive feature functions to later
passes, and using deep network compression to further reduce
computational requirements.  Discriminative segmental cascades allow
us to shift features between passes, giving us greater flexibility in
trading off performance and speed.  In this paper, we further explore
this capability for more efficient segmental cascades.

If we can shift features to later passes, we can reduce pruning time in the first pass.
If $k$ segments are pruned
in the first pass, then the time saved for the second pass
is $O(kd)$ where $d$ is the number of features.
Even though the second pass uses many more features
than the first pass, the decoding time is vastly lower
due to sparseness of the lattices.


Another expense
is feeding forward in neural networks used to compute the feature functions.
For example, in our implementation it takes about $0.33\times$ real-time
to feed forward in a three-layer bidirectional
long short-term memory network (LSTM)
\cite{HochreiterSchmidhuber1997}.

The total recognition time is the sum of the decoding and feed-forward
times.
In this paper, we reduce both.
By making the first pass extremely simple and pushing complex features to later
passes, in combination with frame skipping and high-speed pruning,
we are able to obtain several-fold speedups in recognition and training.
Ultimately, we obtain a segmental system that runs in roughly 1/8 real time and
takes roughly 3 hours to train on TIMIT, not counting feeding forward.
Feeding forward alone takes about 1/6 real time, and training LSTMs takes about 30 hours.
In sum, our proposed system decodes in about 1/3 real time
and takes 33 hours to train with a single four-core CPU.

\section{Segmental cascades}

Before defining segmental models, we first
set up the following notation.
A \term{segment} $e$ is a tuple $(s, t, \ell)$, where
$s$ is the start time, $t$ is the end time, and
$\ell$ is its label.
Let $y$ be a sequence of segments
$((s_1, t_1, \ell_1), \dots, (s_K, t_K, \ell_K))$.
A sequence of connected segments is also called
a path.

Let $\mathcal{X}$ be the set of sequences of input acoustic vectors.
For any $x \in \mathcal{X}$, let $\mathcal{Y}(x)$
be the set of paths that cover $x$.
We call $\mathcal{Y}(x)$ the \term{hypothesis space} of $x$.
Let $E$ be the set of all possible segments.
A \term{segmental model} is a tuple $(\theta, \phi)$,
where $\theta \in \mathbb{R}^d$ are the model parameters and
$\phi: \mathcal{X} \times E \to \mathbb{R}^d$
is a vector of feature functions.
For any segmental model, we can use
it to predict (decode) by finding the best-scoring path:
\begin{equation}
\argmax_{y \in \mathcal{Y}(x)} \:\theta^\top \phi(x, y),
\end{equation}
where 
$\phi(x, y) = \sum_{e \in y} \phi(x, e)$
(here we treat the sequence $y$ as
a set and $e$ enumerates over the segments).

For efficient decoding, we will encode a hypothesis
space $\mathcal{Y}(x)$ as a finite-state transducer (FST).
A finite-state transducer is a tuple $(V, E, I, F, \tail, \head, i, o, w)$,
where $V$ is a set of vertices, $E$ is a set of edges,
$I$ is a set of starting vertices, $F$ is a set of ending
vertices, $\tail: E \to V$ is a function that maps an edge
to its starting vertex, $\head: E \to V$ is a function
that maps an edge to its ending vertex,
$i: E \to \Sigma$ is a function that maps
an edge to an input symbol in the input symbol set $\Sigma$,
$o: E \to \Lambda$ is a function that maps an edge
to an output symbol in the output symbol set $\Lambda$,
and $w: E \to \mathbb{R}$ is a function that maps
an edge to a weight.  We also define $\text{adj}(v) = \{e: \tail(e) = v\}$,
the set of edges that start at vertex $v$. 
We deliberately reuse $E$ in this definition since edges in the FST correspond 
to segments in the hypothesis space. 


A \term{segmental cascade} \cite{TangEtAl2015} is a multi-pass system,
consisting of a sequence of segmental models $A_1, A_2, \dots, A_M$.
Given $x \in \mathcal{X}$, we first train
$A_1 = (\theta_1, \phi_1)$ on $\mathcal{Y}_1 = \mathcal{Y}(x)$,
and use $A_1$ to prune $\mathcal{Y}_1$ 
to produce $\mathcal{Y}_2$.
In general, for any $i$, we train $A_i$ on $\mathcal{Y}_i$
and produce $\mathcal{Y}_{i+1}$.
In the end, we use the final segmental model $A_M$ to
predict by choosing the best scoring path
in $\mathcal{Y}_M$.  The hypothesis
spaces $\mathcal{Y}_2, \mathcal{Y}_3, \dots$
are often called \term{lattices}.

Based on \cite{TangEtAl2014}'s comparison of various losses and costs for
training segmental models, we use hinge loss with overlap cost.  In
the next several sections, we introduce several ingredients that we
explore for speeding up decoding and training. 

\section{Efficiency measure 1:  A simple two-feature first-pass model}


The first efficiency measure we explore is a first pass segmental
model that is as simple as possible, in order to prune the initial
hypothesis space as fast as possible.
In particular, to use as few features as possible in the first pass,
we propose a two-feature segmental model that,
as the name suggests, uses only two features: a label posterior feature defined below,
and a bias.

We assume the existence of a neural network classifier that 
produces
a posterior for each possible label for each frame of acoustic
input.
For any segment $(s, t, \ell)$,  the \term{label posterior feature}
$\phi_p$ is the sum of the log posterior of the label $\ell$ (according to the
frame classifier) over the frames in the segment:
\begin{equation}
  \phi_p(x, (s, t, \ell)) = \sum_{k = s}^t \log h(x)_{k, \ell}
\end{equation}
where $h(x)$ is the sequence of output vectors
of the 
network given $x$,
and $h(x)_{k, \ell}$ is the $\ell$-th vector element 
at time $k$. 

\section{Efficiency measure 2:  Pruning}

One major component in segmental cascades is pruning.
Different pruning methods produce lattices with
different properties.
We consider three
pruning methods, described next, and will experimentally compare
the quality of lattices they produce.

\subsection{Beam pruning}

Beam pruning \cite{SteinbissEtAl1994} is a widely used pruning method, and
it includes many variants with different pruning criteria \cite{Aubert2002}.
For a precise comparison, we describe in detail the beam pruning algorithm we
use here. 

The algorithm keeps track of the maximum partial scores of
paths reaching a vertex $v$, $d(v)$;
the vertices it will explore, $\mathcal{B}$;
and the paths explored so far, $\mathcal{P}$.
The algorithm starts with $\mathcal{B} \gets I, \mathcal{P} \gets \emptyset$,
and $d(i) = 0$ for all $i \in I$.
The vertices are traversed in topological order.
Suppose $v$ is traversed, and $v \in \mathcal{B}$.
We calculate the maximum and minimum score branching out from $v$:
\begin{align}
s_{\max} & = d(v) + \max_{e \in \adj(v)} w(e) \\
s_{\min} & = d(v) + \min_{e \in \adj(v)} w(e)
\end{align}
After computing the maximum and minimum, we use them to compute the
threshold
\begin{align*}
t = \alpha_\text{beam} s_{\max} + (1 - \alpha_\text{beam}) s_{\min},
\end{align*}
where $\alpha_\text{beam}$ controls
the proportion of edges to be pruned.  Edges are pruned locally
while branching out according to the threshold.
Edges that survived pruning are kept in the set
\begin{align*}
\mathcal{S}_v = \{e \in \adj(v): d(v) + w(e) > t\}.
\end{align*}
Paths are expanded as well:
\begin{align*}
\mathcal{P} & \gets \bigcup_{p \in \mathcal{P}} \{pe : e \in \mathcal{S}_v\},
\end{align*}
where $pe$ is the concatenation of $p$ and $e$.
The maximum scores are updated for every $e \in \mathcal{S}_v$, and
the algorithm keeps the vertices that are not pruned and moves on to the
next reachable vertex:
\begin{align*}
d(\head(e)) & \gets \max(d(\head(e)), d(\tail(e)) + w(e)) \\
\mathcal{B} & \gets (\mathcal{B} \setminus \{v\}) \cup \{\head(e) : e \in \mathcal{S}_v\}
\end{align*}
Vertices not reachable from $\mathcal{B}$ are simply ignored.
The set of paths $\mathcal{P}$ comprise the final output lattice.

\subsection{Max-marginal edge pruning}


The \term{max-marginal} \cite{SixtusOrtmanns1999, WeissEtAl2012} of an
edge is defined as the maximum path score among the paths
that pass through $e$:
\begin{equation}
\gamma(e) = \max_{y \ni e} \:\theta^\top \phi(x, y).
\end{equation}
We prune an edge $e$ if its max-marginal $\gamma(e)$
is below a certain threshold.
We use a threshold of the following form
\begin{equation*}
t = \alpha_\text{edge} \max_{e \in E} \gamma(e)
    + (1 - \alpha_\text{edge}) \frac{1}{|E|} \sum_{e \in E} \gamma(e),
\end{equation*}
where $\alpha_\text{edge}$ controls
the proportion of edges that are pruned.
Max-marginal edge pruning has a guarantee that
all paths left unpruned have higher scores than the threshold,
while beam pruning has none.  However, for max-marginal
edge pruning, we need to enumerate the edges at least
once, while for beam pruning, we are able to ignore
some of the edges without even accessing them.

\subsection{Max-marginal vertex pruning}

The \term{max-marginal} of a vertex is defined as
\begin{equation}
\gamma(v) = \max_{y \ni v} \theta^\top \phi(x, y).
\end{equation}
In words, it is the maximum path score among the paths that pass through $v$.
We prune a vertex $v$ if its max-marginal $\gamma(v)$ is below
a certain threshold.
Similarly to edge pruning, we use the threshold
\begin{equation*}
t = \alpha_\text{vertex} \max_{v \in V} \gamma(v)
    + (1 - \alpha_\text{vertex}) \frac{1}{|V|} \sum_{v \in V} \gamma(v),
\end{equation*}
where $\alpha_\text{vertex}$ controls the aggressiveness of the pruning.

Because of the way we construct FSTs from hypothesis spaces,
a vertex corresponds to the time of the boundary 
between its
two neighboring segments.  When a vertex is pruned,
it indicates that the corresponding time point
is unlikely to be a true boundary between segments.

\section{Efficiency measure 3:  Subsampling}


In our segmental models, we use deep bidirectional LSTMs trained
separately with
log loss
to produce frame posteriors
$h_1, \dots, h_T$ given input vectors $x_1, \dots, x_T$ and frame
label hypotheses $y_1, \dots, y_T$.  LSTMs are increasingly popular
for speech recognition and, in recent work, have obtained excellent performance as feature
generators for segmental models~\cite{LuEtAl2016}.
Not surprisingly, feeding forward through the networks contributes
a large part of the total computation time.
Following \cite{VanhouckeEtAl2013, MiaoEtAl2015}, we consider dropping half of the
frames for any given utterance in order to save time on feeding
forward.
Specifically, we only use $x_2, x_4, \dots, x_{T - 2}, x_T$
to feed forward through the deep LSTM and generate $h_2, h_4, \dots,
h_{T - 2}, h_T$ (assuming $T$ is even, without loss of generality).
We then copy each even-indexed output to its previous frame, i.e., $h_{i-1} = h_i$
for $i = 2, 4, \dots, T-2, T$.
During training, the log 
loss is calculated over all frames
and propagated back.  Specifically,
let $E_i$ be the log loss at frame $i$, and $E = \sum_{i=1}^T E_i$.
The gradient of $h_i$ is the sum of the gradients from the current frame
and the copied frame.
Dropping even-indexed frames is similar to dropping odd-indexed frames,
except the outputs are copied from $h_i$ to $h_{i+1}$ for $i = 1, 3, \dots, T-1$.

\section{Experiments}

We test our proposed 
efficiency measures for the task of phonetic recognition
on the TIMIT dataset.  An utterance in TIMIT is
about 3 seconds on average, so absolute wall-clock time corresponding to
1 times real-time is 3 seconds per utterance. 

For the following experiments, the running time is measured on a machine
with a Core i7-2600 3.4 GHz quad-core CPU.
All pruning and decoding experiments are done with a single thread,
and all neural network feeding forward and back-propagating
are done with four threads.

\subsection{LSTM frame classification}

Most of our segmental features are computed from the outputs of an
LSTM, so we first explore its performance.  We build a frame classifier by stacking 3 layers of bidirectional LSTM.
The cell and hidden vectors have 256 units.  We train the frame classifier
with frame-wise log-loss and optimize with AdaGrad~\cite{DuchiEtAl2011} 
with mini-batch size of 1 and step size 0.01 for 30 epochs.
We choose the best-performing model on the development set (early stopping).

\begin{figure}
\begin{center}
\includegraphics[width=1.5in]{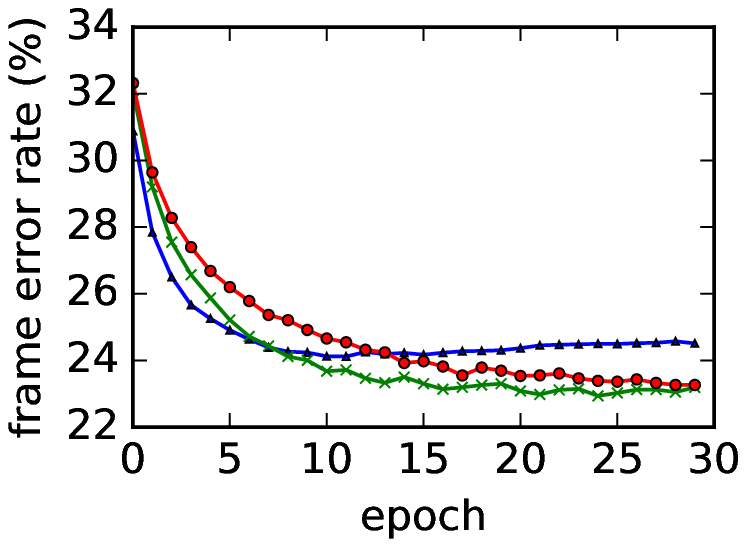}
\includegraphics[width=1.5in]{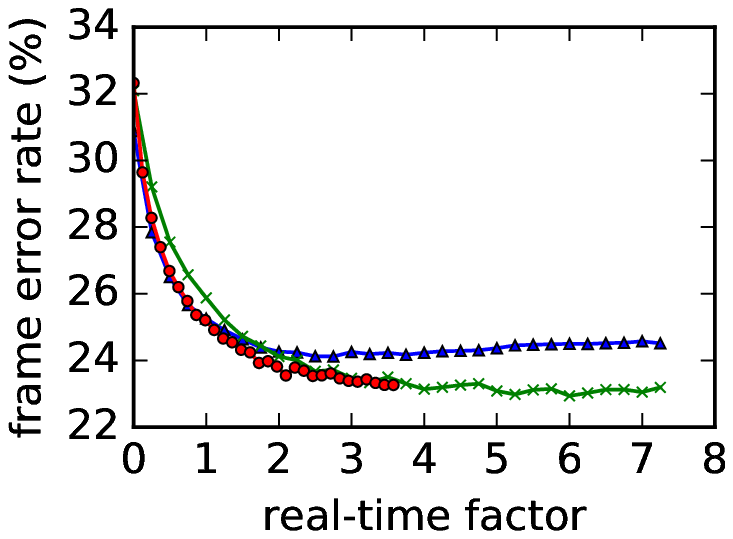}
\caption[]{Development set frame error rates vs. number of epochs (left)
    and vs. training real-time factor (time spent training / duration of training set,
    right).
    \label{fig:frame}
    \begin{tikzpicture}[baseline=-\the\dimexpr\fontdimen22\textfont2\relax]
    \node [right] at (0.2, 0) {vanilla};
    \draw[blue] (-0.2, 0) -- (0, 0);
    \draw[blue] (0, 0) -- (0.2, 0);
    \filldraw[fill=blue,draw=black] (-0.07, -0.04) -- (0.07, -0.04)
        -- (0, 0.08) -- cycle;
    \end{tikzpicture}
    \begin{tikzpicture}[baseline=-\the\dimexpr\fontdimen22\textfont2\relax]
    \node [right] at (0.2, 0) {dropout};
    \draw[black!50!green] (-0.2, 0) -- (0, 0);
    \draw[black!50!green] (0, 0) -- (0.2, 0);
    \draw[black!50!green] (-0.07, 0.07) -- (0.07, -0.07);
    \draw[black!50!green] (-0.07, -0.07) -- (0.07, 0.07);
    \end{tikzpicture}
    \begin{tikzpicture}[baseline=-\the\dimexpr\fontdimen22\textfont2\relax]
    \node [right] at (0.2, 0) {subsampling+dropout};
    \draw[red] (-0.2, 0) -- (0, 0);
    \draw[red] (0, 0) -- (0.2, 0);
    \filldraw[fill=red,draw=black,radius=0.06] (0, 0) circle;
    \end{tikzpicture}
    }
\end{center}
\end{figure}

We compare training the LSTM with and without dropout.
Following \cite{ZarembaEtAl2014}, we use dropout on the input vectors
of the second and third layer.
The results are shown in Figure~\ref{fig:frame}.
Without dropout, we observe that the frame error rates
on the development set slightly increase toward the end, while
with dropout, the frame error rates improve by about 1\%
absolute.
\klcomment{skip this par since LSTM dropout isn't new or a part of the contribution?}

Next we consider the effect of frame subsampling.
When training LSTMs with subsampling, we alternate
between dropping even- and odd-numbered frames every other epoch.
Other training hyper-parameters remain the same.
We observe that with subsampling the model converges
more slowly than without, in terms of number of epochs.
However, by the end of epoch 30, there is almost no loss
in frame error rates when we drop half of the frames.
Considering the more important measure of training time rather than number of epochs,
the LSTMs with frame subsampling
converge twice as fast as those without subsampling.
For the remaining experiments, we use the log posteriors at each frame
of the subsampled LSTM outputs as the
inputs to
the segmental models.

\begin{table*}
\caption{Real-time factors for decoding per sample
    and hours for training the system.
\label{tbl:train}}
\begin{center}
\begin{tabular}{l||p{0.5cm}p{0.5cm}p{0.5cm}p{1.3cm}|p{1.1cm}p{1cm}||p{0.5cm}p{0.5cm}p{0.5cm}p{1.3cm}|p{1.1cm}p{1cm}}
          & \multicolumn{4}{l}{decoding (RT)} & & & \multicolumn{2}{l}{training (hours)} & \\
\hline
           & 1st pass & 2nd pass & 3rd pass & total decoding & feeding forward & total overall  & 1st pass & 2nd pass & 3rd pass & total training & feeding forward & total overall \\
\hline                                                                                                                                                              
baseline   & 0.33     & 0.01     &          & 0.34           & 0.33            & 0.67           & 49.5     & 0.6      &          & 50.1           & 59.4            & 109.5  \\
proposed   & 0.11     & 0.02     & 0.01     & 0.14           & 0.17            & 0.31           & 1.0      & 1.2      & 0.6      & 2.8            & 29.7            & 32.5  \\
\end{tabular}
\end{center}
\end{table*}

\subsection{Segmental cascade experiments}

Our baseline system is a discriminative segmental model $R$ based on \cite{TangEtAl2015}, which is a
first-pass segmental model using the following segment features:
log posterior averages, log posterior samples,
log posteriors at the boundaries,
segment length indicators, and bias.
See \cite{TangEtAl2015} for more complete descriptions of the
features.
All features are lexicalized with label unigrams.
The baseline system is trained by optimizing hinge loss 
with AdaGrad using mini-batch size 1, step size 0.1,
and early stopping for 50 epochs.

Recall that the proposed two-feature first-pass system is a segmental model $A_1$
with just the label posterior and bias features.
We use hinge loss optimized with AdaGrad with mini-batch size 1 and step size 1.
None of the features are lexicalized.
Since we only have two features, learning
converges very quickly, in only three epochs.  We take the model from
the third epoch to produce lattices for subsequent passes in the 
cascade.

\begin{figure}
\begin{center}
\includegraphics[width=1.4in]{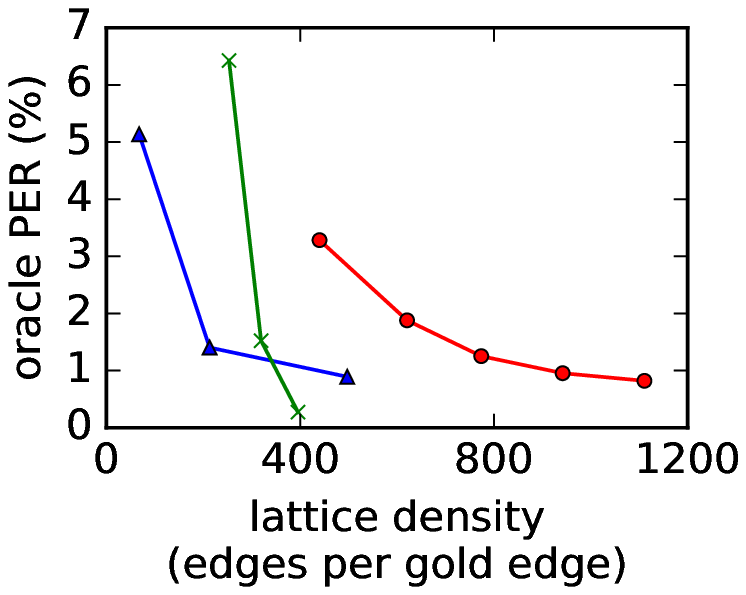}
\includegraphics[width=1.5in]{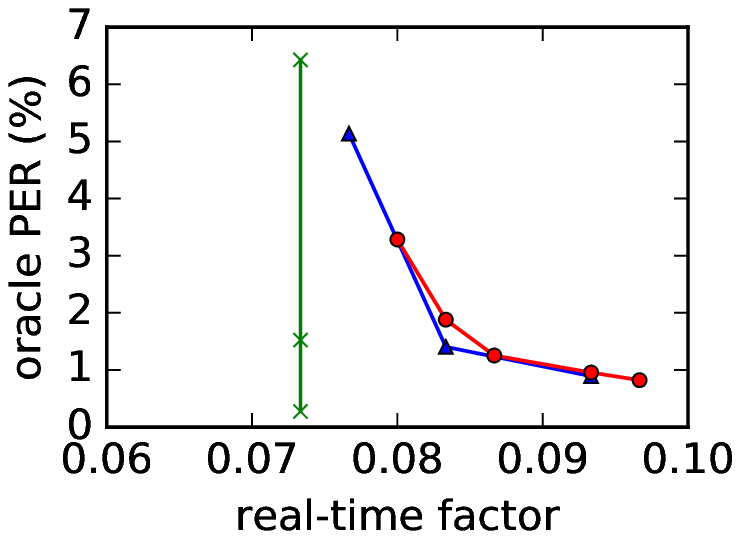}
\caption[]{
    \emph{Left:} Densities versus oracle error rates of lattices
        generated by different pruning methods.
    \emph{Right:} Time spent on generating the lattices with different
    pruning methods. 
    \label{fig:pruning}
    \begin{tikzpicture}[baseline=-\the\dimexpr\fontdimen22\textfont2\relax]
    \node [right] at (0.2, 0) {edge pruning};
    \draw[blue] (-0.2, 0) -- (0, 0);
    \draw[blue] (0, 0) -- (0.2, 0);
    \filldraw[fill=blue,draw=black] (-0.07, -0.04) -- (0.07, -0.04)
        -- (0, 0.08) -- cycle;
    \end{tikzpicture}
    \begin{tikzpicture}[baseline=-\the\dimexpr\fontdimen22\textfont2\relax]
    \node [right] at (0.2, 0) {vertex pruning};
    \draw[black!50!green] (-0.2, 0) -- (0, 0);
    \draw[black!50!green] (0, 0) -- (0.2, 0);
    \draw[black!50!green] (-0.07, 0.07) -- (0.07, -0.07);
    \draw[black!50!green] (-0.07, -0.07) -- (0.07, 0.07);
    \end{tikzpicture}
    \begin{tikzpicture}[baseline=-\the\dimexpr\fontdimen22\textfont2\relax]
    \node [right] at (0.2, 0) {beam pruning};
    \draw[red] (-0.2, 0) -- (0, 0);
    \draw[red] (0, 0) -- (0.2, 0);
    \filldraw[fill=red,draw=black,radius=0.06] (0, 0) circle;
    \end{tikzpicture}
    }
\end{center}
\end{figure}

Given $A_1$, we compare lattices generated by $A_1$
with different pruning methods.
We consider $\alpha_\text{edge} \in \{0.8, 0.85, 0.9\}$ for edge pruning,
$\alpha_\text{vertex} \in \{0.925, 0.95, 0.975\}$ for vertex pruning,
and $\alpha_\text{beam} \in \{0.93, 0.94, 0.95, 0.96\}$ for beam pruning.
The results are shown in Figure~\ref{fig:pruning}.
We observe that edge pruning generally gives the best density-oracle
error rate tradeoff.
Beam search is inferior in terms of speed and lattice quality.

We also measure real-time factors for different pruning methods,
shown in Figure~\ref{fig:pruning}.
We can see that all pruning methods can obtain oracle error rates
less than 2\% within 0.1 $\times$ real-time.
No single pruning method is significantly faster than the
others.  Intuitively, while max-marginal pruning requires more computation per
edge, it makes more informed decisions.  The bottom line is that
max-marginal pruning produces less-dense lattices with the same
performance in the same amount of time.

Based on the above experiments, we use lattices produced by edge pruning
with $\alpha_\text{edge} = 0.85$ for the remaining experiments,
because the lattices are sparse and have only 1.4\%
oracle error rate.
We train a second-pass
model $A_2$ with the same features as the baseline system $R$
except that we add a ``lattice score'' feature corresponding to
the segment score given by the two-feature system $A_1$.
Hinge loss is optimized with AdaGrad with mini-batch size 1, step size 0.1,
and early stopping for 20 epochs.

\begin{figure}
\begin{center}
\includegraphics[width=2in]{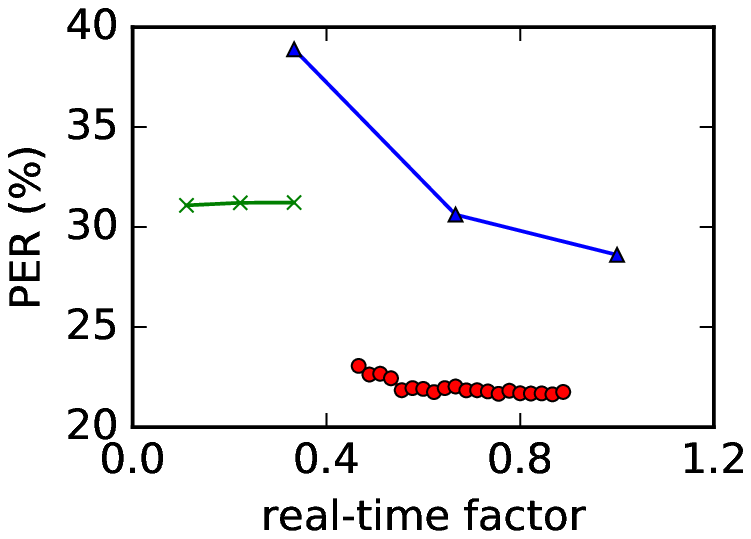}
\caption[]{Learning curve of the proposed two-pass system compared with
    the baseline system.
    The time gap between the first pass and the second pass is the
    time spent on pruning.
    \label{fig:learning}
    \begin{tikzpicture}[baseline=-\the\dimexpr\fontdimen22\textfont2\relax]
    \node [right] at (0.2, 0) {baseline 1st pass ($R$)};
    \draw[blue] (-0.2, 0) -- (0, 0);
    \draw[blue] (0, 0) -- (0.2, 0);
    \filldraw[fill=blue,draw=black] (-0.07, -0.04) -- (0.07, -0.04)
        -- (0, 0.08) -- cycle;
    \end{tikzpicture}
    \begin{tikzpicture}[baseline=-\the\dimexpr\fontdimen22\textfont2\relax]
    \node [right] at (0.2, 0) {proposed 1st pass ($A_1$)};
    \draw[black!50!green] (-0.2, 0) -- (0, 0);
    \draw[black!50!green] (0, 0) -- (0.2, 0);
    \draw[black!50!green] (-0.07, 0.07) -- (0.07, -0.07);
    \draw[black!50!green] (-0.07, -0.07) -- (0.07, 0.07);
    \end{tikzpicture}
    \begin{tikzpicture}[baseline=-\the\dimexpr\fontdimen22\textfont2\relax]
    \node [right] at (0.2, 0) {proposed 2nd pass ($A_2$)};
    \draw[red] (-0.2, 0) -- (0, 0);
    \draw[red] (0, 0) -- (0.2, 0);
    \filldraw[fill=red,draw=black,radius=0.06] (0, 0) circle;
    \end{tikzpicture}
    }
\end{center}
\end{figure}

The learning curve comparing $R$ and $A_1$ followed by $A_2$
is shown in Figure~\ref{fig:learning}.
We observe that the learning time per epoch of the two-feature system $A_1$
is only one-third of the baseline system $R$.
We also observe that training of $A_2$ converges faster
than training $R$, despite the fact that they use almost identical
feature functions.
The baseline system achieves the best result at epoch 49.
In contrast, the two-pass system is done
before the baseline even finishes the third epoch.

Following \cite{TangEtAl2015}, given the first-pass baseline,
we apply max-marginal
edge pruning to produce lattices for the second-pass
baseline with $\alpha_\text{edge}=0.8$.  The second-pass baseline features
are the lattice score from the first-pass baseline, a bigram
language model score, first-order length indicators,
and a bias.  Hinge loss is optimized with AdaGrad with mini batch size 1, step size 0.01, and early stopping for 20 epochs.
For the proposed system, we produce lattices with
edge pruning and $\alpha_\text{edge}=0.3$ for the third-pass system.
We use the same set of features and hyper-parameters
as the second-pass baseline for the third pass.

Phone error rates of all
passes are shown in Table~\ref{tbl:per}.
First, if we compare the one-pass baseline
with the proposed two-pass system,
our system is on par with the baseline.
Second, we observe a healthy improvement by just adding
the bigram language model score to the second-pass baseline. 
The improvement for our third-pass system is small
but brings our final performance to within 0.4 of the baseline second pass.

\begin{table}
  \caption{Phone error rates (\%) of proposed and baseline systems.
    \label{tbl:per}}
\begin{center}
\begin{tabular}{ll|lll}
         &       & 1st pass & 2nd pass & 3rd pass \\
\hline
baseline & dev   & 21.9     & 21.0     &      \\
         & test  & 24.0     & 23.0     &      \\
\hline
proposed & dev   & 33.6     & 21.5     & 21.3 \\
         & test  &          & 23.7     & 23.4 \\
\end{tabular}
\end{center}
\end{table}

Next we report on the speedups in training
and decoding obtained with our proposed approach.
Table~\ref{tbl:train} shows the real-time factors
for decoding with the baseline and proposed systems.
In terms of decoding time alone, we achieve
a 2.4 time speedup compared to the baseline.
If the time of feeding forward LSTMs is included, then
our proposed system is two times faster
than the baseline.

Table~\ref{tbl:train} shows the times needed
to train a system to get to the performance in Table~\ref{tbl:per}.
The speedup mostly comes from the fast convergence
of the first pass.
In terms of training the segmental models alone,
we achieve an 18.0-fold speedup.
If the time to train the LSTMs is included, then
we obtain a 3.4-fold speedup
compared to the baseline.

To summarize some of the above results:  With a combination of the first-pass two-feature
system and edge pruning, we prune 95\% of
the segments in the first-pass hypothesis space,
leading to significant speedup in both decoding and training.
The feed-forward time for our LSTMs is halved through frame subsampling.
In the end, with a single four-core CPU, we achieve
0.31 times real-time decoding
including feeding forward,
which is 2.2 times faster than the baseline, 
and 32.5 hours in total to obtain our final model
including LSTM training, which is
3.4 times faster than the baseline.
Excluding the LSTMs, the segmental model decoding alone is 2.4 times faster than the baseline,
and training the segmental models alone is 18 times faster than the baseline.

\section{Conclusion}

We have studied efficiency improvements to segmental
speech recognition structured as a discriminative segmental cascade.
Segmental cascades allow us to push features around between different
passes to optimize speed and performance.  We have taken
advantage of this and proposed an extremely simple first-pass segmental model
with just two features, a label posterior and a bias, which is
intuitively like a segmental analogue of a typical hybrid model.
We have also compared pruning approaches and find that max-marginal edge pruning is the most effective
in terms of time, lattice density, and oracle error rates.
With the combination of these measures and frame subsampling for our
input LSTMs, we obtain large gains in decoding and training speeds.
With decoding and training now being much faster, future work will
explore even richer feature functions and larger-scale tasks such as word recognition.

\section{Acknowledgements}
This research was supported by a Google faculty research award and NSF
grant IIS-1433485.  The opinions expressed in this work are those of
the authors and do not necessarily reflect the views of the funding
agency. 
The GPUs used for this research were donated by NVIDIA.

\clearpage
\eightpt
\bibliographystyle{IEEEtran}
\bibliography{prune}

\end{document}